\documentclass[10pt,twocolumn,letterpaper]{article}

\usepackage{cvpr}              

\usepackage{graphicx}
\usepackage{amsmath}
\usepackage{amssymb}
\usepackage{booktabs}

\usepackage[pagebackref,breaklinks,colorlinks]{hyperref}

\usepackage{multirow}
\usepackage{array}
\usepackage{booktabs} 
\usepackage{subcaption}
\usepackage{sidecap}
\usepackage{textcomp}
\usepackage{makecell}
\usepackage{amsmath}
\usepackage{xcolor}         
\usepackage[capitalize]{cleveref}
\usepackage{amsfonts}       
\usepackage{nicefrac}       
 \usepackage{soul}
  \usepackage{arydshln}

\usepackage[capitalize]{cleveref}
\crefname{section}{Sec.}{Secs.}
\Crefname{section}{Section}{Sections}
\Crefname{table}{Table}{Tables}
\crefname{table}{Tab.}{Tabs.}

\def\cvprPaperID{10799} 
\def\confName{CVPR}
\def\confYear{2023}

\begin{document}

\title{Open-Set Representation Learning through Combinatorial Embedding}

\author{%
  Geeho Kim$^1$ \qquad Junoh Kang$^1$ \qquad Bohyung Han$^{1, 2}$\\
Computer Vision Laboratory, ECE$^1$ \& IPAI$^{2}$, Seoul National University \\
  \texttt{\{snow1234, junoh.kang, bhhan\}@snu.ac.kr}}
  \maketitle


\begin{abstract} 

Visual recognition tasks are often limited to dealing with a small subset of classes simply because the labels for the remaining classes are unavailable. We are interested in identifying novel concepts in a dataset through representation learning based on both labeled and unlabeled examples, and extending the horizon of recognition to both known and novel classes.
To address this challenging task, we propose a combinatorial learning approach, which naturally clusters the examples in unseen classes using the compositional knowledge given by multiple supervised meta-classifiers on heterogeneous label spaces.
The representations given by the combinatorial embedding are made more robust by unsupervised pairwise relation learning.
The proposed algorithm discovers novel concepts via a joint optimization for enhancing the discrimitiveness of unseen classes as well as learning the representations of known classes generalizable to novel ones.
Our extensive experiments demonstrate remarkable performance gains by the proposed approach on public datasets for image retrieval and image categorization with novel class discovery.

\end{abstract} 


\section{Introduction} 
\label{sec:introduction}
Despite the remarkable success of machine learning fueled by deep neural networks, existing frameworks still have critical limitations in an open-world setting, where some categories are not defined a priori and the labels for some classes are missing.
Although there have been a growing number of works that identify new classes in unlabeled data given a set of labeled examples~\cite{hsu2018learning, cao2021open, hsu2019multiclass, brbic2020mars, Han2019learning, han2020automatically}, they often assume that all the unlabeled examples belong to unseen classes and/or the number of novel classes is known in advance, which makes their problem settings unrealistic.

To address the limitations, this paper introduces an algorithm applicable to a more realistic setting.
We aim to discover and learn the representations of unseen categories without any prior information or supervision about novel classes, where unlabeled data may contain examples in both seen and unseen classes.
This task requires the model to be able to effectively identify unseen classes while preserving the information of previously seen classes.
Our problem setting is more challenging than the case where the unlabeled data only consist of unseen classes because we have to solve an additional problem, predicting the membership of unlabeled examples between seen and unseen classes.

\begin{figure*}[t]
\centering
\begin{subfigure}[b]{0.3\linewidth}
\centering
\includegraphics[width=1\linewidth]{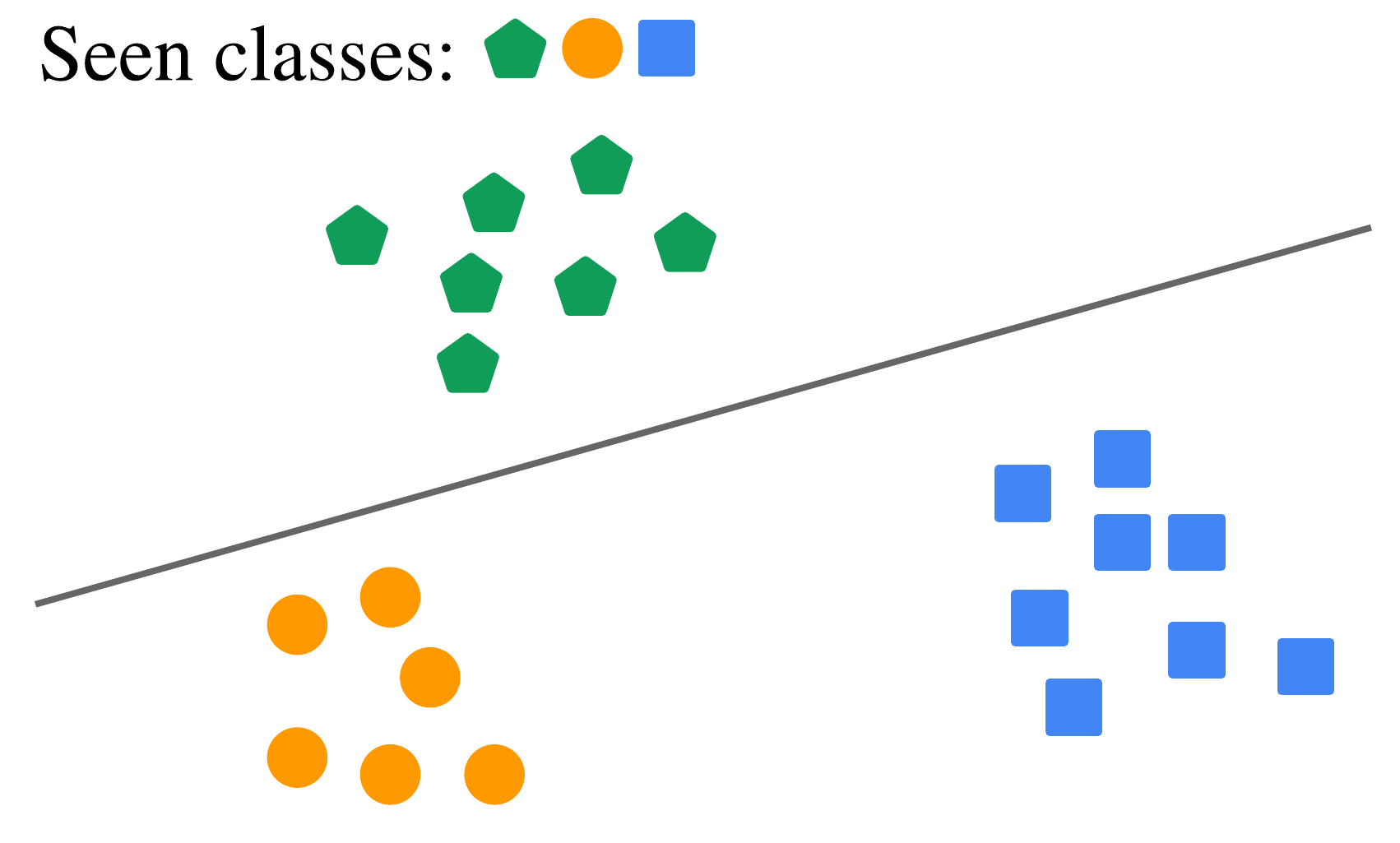}
\vspace{-0.4cm}
\caption{One meta-class set}
\end{subfigure}
\hspace{0.5cm}
\begin{subfigure}[b]{0.3\linewidth}
\centering
\includegraphics[width=1\linewidth]{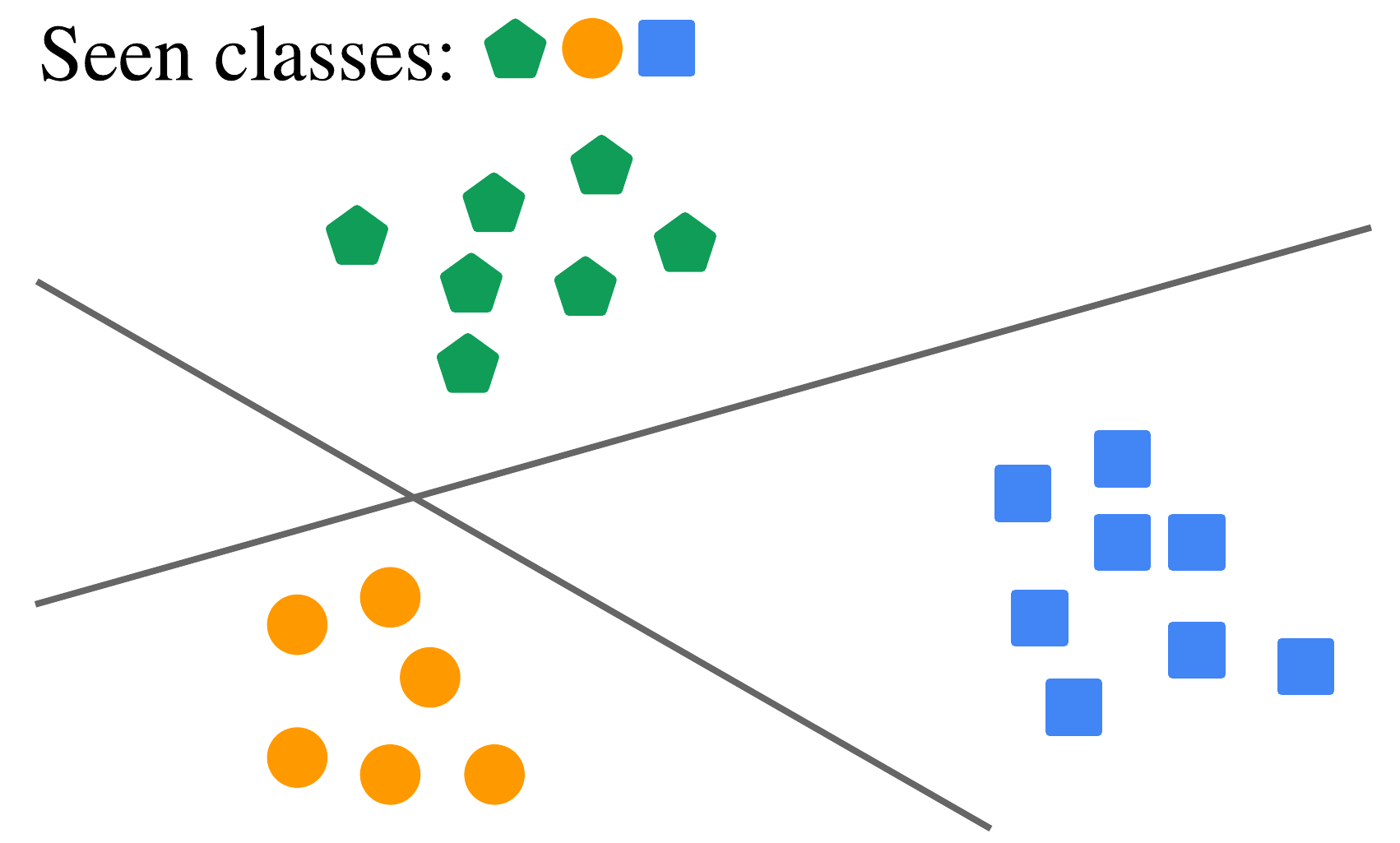}
\vspace{-0.4cm}
\caption{Two meta-class sets}
\end{subfigure}
\hspace{0.5cm}
\begin{subfigure}[b]{0.3\linewidth}
\centering
\includegraphics[width=1\linewidth]{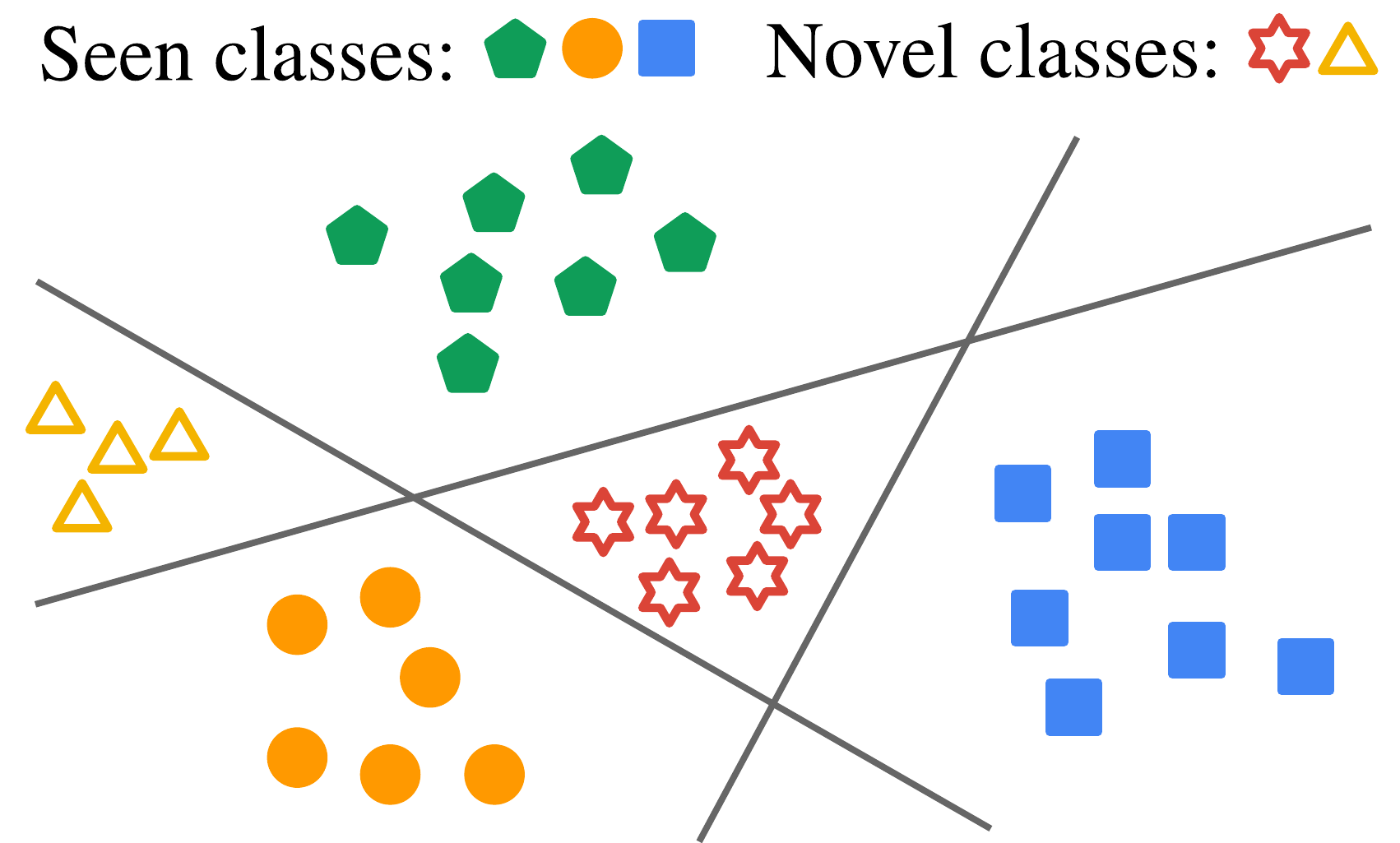}
\vspace{-0.4cm}
\caption{Three meta-class sets}
\end{subfigure}
\caption{
Conceptual illustration of decision boundaries (black solid lines) given by combinatorial classification with three seen classes, where three binary meta-classifiers are added one-by-one from (a) to (c).
Unlike the standard classifier that creates decision boundaries for seen classes only, the combinatorial classification based on multiple coarse-grained classifiers creates and reserves partitions, which are distinct from those of seen classes, potentially corresponding to novel concepts.}
\label{fig:main_concept}
\end{figure*}

We propose a representation learning approach based on the concept of combinatorial classification~\cite{seo2019combinatorial}, where the examples in unseen categories are identified by the composition of multiple meta-classifiers. 
Figure~\ref{fig:main_concept}~illustrates the main idea of our \emph{combinatorial embedding} framework, which forms partitions for novel classes via a combination of multiple classifiers for the meta-classes involving several constituent base classes.
Images in the same meta-class potentially have common attributes that are helpful for knowledge transfer to novel classes, and we learn the representations of the images by the proposed combinatorial embedding.
The learned representations via the combinatorial embedding become even stronger by unsupervised pairwise relation learning, which is effective to identify novel classes.

Our main contributions are summarized as follows.
 \begin{itemize} 
	\item[$\bullet$] 
	 We propose a novel combinatorial learning framework, which embeds the examples in both seen and novel classes effectively by the composition of the knowledge learned from multiple heterogeneous meta-class classifiers.
	\item[$\bullet$] We introduce an unsupervised learning approach to define pairwise relations, especially semantic structure between labeled and unlabeled examples, which further improves the quality of the representations given by combinatorial embedding.
	\item[$\bullet$] We demonstrate the outstanding performance of our model in the presence of novel classes through extensive evaluations on image retrieval and image categorization with novel class discovery benchmarks.
\end{itemize}

In the rest of this paper, we first review related works in Section~\ref{sec:related} and discusses our main algorithm in Section~\ref{sec:proposed}. 
Section~\ref{sec:experiments} presents our experimental results and Section~\ref{sec:conclusion} concludes this paper.


\section{Related work} 
\label{sec:related}

This section first introduces recent approaches for open-set recognition, and then discusses several related methods to combinatorial learning.

\subsection{Learning in Open-Set Setting}
Departing from the closed world, a number of works recently consider the open-set setting, where novel classes appear during testing~\cite{scheirer2012toward, geng2020recent, scheirer2014probability, scholkopf2001estimating, bendale2016towards, yoshihashi2019classification, vareto2017towards, junior2017nearest, hendrycks17baseline}.
Early researches mainly focus on detecting out-of-distribution examples by learning binary classifiers~\cite{scheirer2012toward, scholkopf2001estimating, vareto2017towards, hendrycks17baseline}, or classifying the knowns while rejecting the unknowns~\cite{scheirer2014probability, bendale2016towards, yoshihashi2019classification, junior2017nearest}.
However, these approaches have significant challenges in distinguishing semantics between unseen classes; although some methods sidestep the issue by assigning rejected instances to new categories~\cite{bendale2015towards, shu2020p, shu2018odn}, they require human intervention to annotate the rejected examples and consequently suffer from weak scalability.

To mitigate such limitations, transfer learning approaches have been proposed to model semantics between unseen classes.
Using the representations learned from labeled data, the methods in this category perform clustering with unlabeled examples based on similarity prediction models~\cite{hsu2018learning, hsu2019multiclass}, ranking statistics~\cite{han2020automatically}, and modified deep embedded clustering~\cite{Han2019learning} to capture their similarity and discrepancy.
However, these approaches have two critical limitations.
First, the problem settings are unrealistic because they assume that all unlabeled examples belong to unseen classes, and the number of novel classes is known in advance.
Second, their main goal is to learn the representations of novel classes, which results in information loss about seen classes.
Recent works~\cite{cao2021open,vaze2022generalized} generalize the problem setting, where the unlabeled instances may come from both seen and novel classes.
Cao~\etal~\cite{cao2021open} revise the standard cross-entropy loss with an adaptive margin to prevent the model from being biased towards the seen classes while Vaze~\etal~\cite{vaze2022generalized} employ two contrastive losses to pretrain the representations and adopt $k$-means++ clustering~\cite{arthur2006k} for evaluation.
However, these approaches still require prior information about the number of novel classes or computationally expensive modules to estimate the number of novel classes.
On the contrary, we do not use any information about the number of novel classes for training because we discover novel categories based on the outputs from the meta-classifiers.

On the other hand, several hashing techniques~\cite{zhang2017ssdh, yan2017semi, jin2020ssah, jang2020generalized} learn approximated embeddings for image retrieval with both labeled and unlabeled data, which is generalizable to the examples in unseen classes.
They focus on reducing quantization distortion in hash function by either entropy minimization~\cite{zhang2017ssdh, jang2020generalized} or consistency regularization~\cite{yan2017semi, jin2020ssah}.

\subsection{Combinatorial Learning}

Combinatorial learning framework reconstructs the solution space by the composition of the solutions from multiple heterogeneous tasks and there are several related approaches in this regard.
Seo~\etal~\cite{seo2018cplanet}~formulate the image geolocalization problem as a classification task by combining multiple coarse-grained classifiers to reduce data deficiency and poor prediction granularity.
A similar concept has been employed to learn noise-resistant classifiers~\cite{seo2019combinatorial} or recognize out-of-distribution examples~\cite{vareto2017towards}.
Xuan~\etal~\cite{xuan2018deep}~concatenate multiple representations learned on multiple class sets for metric learning.

Product quantization~\cite{jegou2010product,ge2013optimized}, which is also related to combinatorial learning, constructs a large number of quantized regions given by a combination of subspace encodings to improve the performance of hash functions in an unsupervised manner.
This approach is extended to learning quantization tables using image labels~\cite{jang2020generalized, klein2019end, yu2018product}.
However, they do not provide direct supervision for quantization but optimize the representation via the final classification loss, making the learned model suboptimal.

While all of these approaches are not studied in the presence of unlabeled examples during training except GPQ~\cite{jang2020generalized}, the proposed algorithm leverages the composition of output representations for capturing the semantics of unlabeled data, which belong to either known or novel classes.
Also, contrary to~\cite{jang2020generalized, klein2019end, yu2018product}, the proposed combinatorial embedding learns the representation with explicit supervision in the form of diverse meta-class labels and obtains a better embedding model for novel classes.


\section{Proposed Approach}
\label{sec:proposed} 
Suppose that we are given a labeled dataset, $\mathcal{D}_l = \{(x_i,y_i)\}_{i=1}^{N_l}$, where $x_i \in \mathbb{R}^d$ denotes an input example and $y_{i} \in \mathcal{C}_l = \{c_1,\dots, c_K\}$ is its class label, as well as an unlabeled dataset $\mathcal{D}_u = \{(x_i)\}_{i=1}^{N_u}$ for training.
Let $\mathcal{C}_l$ and $\mathcal{C}_u$ be the ground-truth class sets of the labeled and unlabeled data, respectively, where $\mathcal{C}_l \cap \mathcal{C}_u \ne \emptyset$ and $\mathcal{C}_l \ne \mathcal{C}_u$. 
We denote the novel class set by ${\mathcal{C}}_n = \mathcal{C}_u \backslash \mathcal{C}_l$. 
Our goal is to learn an unified model that is effective to represent novel classes as well as known ones by taking advantage of semantic relations across the two kinds of classes.

To this end, we propose a supervised combinatorial embedding approach and two unsupervised pairwise learning techniques.
For combinatorial embedding, we first construct multiple heterogeneous meta-class sets, each of which is obtained from a unique partition of the base classes.
We then obtain the combinatorial embedding vector of a base class by concatenating meta-class embeddings learned from the classifiers over the individual meta-class sets.
Along with the supervised learning objective, we also perform unsupervised learning based on contrastive loss and consistency regularization for understanding pairwise relations of both seen and unseen classes.

\subsection{Supervised Combinatorial Embedding}
\label{sub:supervised}

The main idea of the supervised combinatorial embedding is to learn the general representations, which embed known and novel classes in a discriminative way, using a composition of multiple heterogeneous coarse-grained classifiers corresponding to meta-class sets.
Formally, if we are given $M$ coarse-grained classifiers $f^1, f^2, \dots, f^M$, defined over meta-class sets as
\begin{align}
\begin{array}{c}
f^1 : x \in \mathbb{R}^d \rightarrow y \in \mathcal{C}^1 = \{c^1_1, \dots, c^1_{K_1} \} \\
\vdots \\
f^M : x \in \mathbb{R}^d \rightarrow y \in \mathcal{C}^M = \{c^M_1, \dots, c^M_{K_M} \},
\end{array}
\label{eq:meta_classes}
\end{align}
we obtain a fine-grained combinatorial classifier $f \equiv f^1 \times f^2 \times \cdots \times f^M$, which is given by
\begin{align}
f : x \in \mathbb{R}^d \rightarrow y \in \mathcal{C}^1 \times \cdots \times \mathcal{C}^M.
\label{eq:combinatorial_classification}
\end{align}

We first construct $M$ distinct partitions, denoted by $\mathcal{C}^m$ ($m = 1, \dots, M$).
Each partition is referred to as a meta-class set, which has $K_m (\ll K)$ meta-classes,~\ie~$\mathcal{C}^m = \{ c^m_{1}, \dots, c^m_{K_m} \}$, and each meta-class is typically constructed by a union of multiple base classes.
Let an input image $x \in \mathbb{R}^d$ be mapped to a vector ${z} \in \mathbb{R}^{d_1}$ by a feature extractor $f_\theta(\cdot)$, \ie ${z} = f_\theta (x)$.
The feature vector $z$ is partitioned to $M$ distinct subvectors, $z^1, \dots, z^M$ ($z^m \in \mathbb{R}^{d_2}, d_1 = M d_2$), which are feature vectors for learning the meta-classifiers for the corresponding meta-class sets. 
We estimate the embedding of each base class based on the meta-class embeddings in the meta-classifiers.
Specifically, we construct $M$ embedding heads with weight matrix $\Theta = \{ \Theta^1, \cdots, \Theta^M \}$ ($\Theta^m \in \mathbb{R}^{d_2 \times K_m}$), and each head corresponds to a classifier for a meta-class set $\mathcal{C}^m$ whose parameters consist of the prototypes of the meta-classes, denoted by ${\Theta}^m = \big[ \theta^m_1, \cdots, \theta^m_{K_{m}} \big]$ ($\mathbf{\theta}^m_k \in \mathbb{R}^{d_2}$).

The combinatorial embedding of a base class is given by a concatenation of the meta-class embeddings, which is formally given by $\pi(z; \Theta) = \big[\Phi(z^1, \Theta^1), \cdots, \Phi(z^M, \Theta^M)\big] \in \mathbb{R}^{d_2M}$. 
Note that $\Phi(\cdot, \cdot)$ performs the soft assignment~\cite{yu2018product} of $z^{m}$ to individual meta-classes to enable backpropagation as
\begin{equation}
\Phi(z^m, {\Theta}^m)=\sum_{i=1}^{K_m} \frac{\exp \left( \lambda (z^{m} \cdot \mathbf{\theta}^{m}_i) \right)}{\sum_{j=1}^{K_m} \exp \left( \lambda (z^{m} \cdot \mathbf{\theta}^{m}_{j}) \right) } \mathbf{\theta}^{m}_{i},
\end{equation}
where $\lambda$ is a sufficiently large scaling factor to approximate the function to a discrete argmax function.
Feature vectors and embedding weights are $\ell_2$-normalized before the inner product to use cosine similarity as the distance metric.
Note that the proposed embedding function enables us to characterize the semantics of unlabeled samples using their embeddings.
For instance, supposing that an example in a novel class has the same meta-class label as those in some known classes with a meta-class set while having a different one in another meta-class set, we can compute the unique embedding of the novel class with respect to those of the seen classes.

Using all the labeled examples, for which meta-class labels are also available, our model learns the representations based on the meta-class labels using the normalized softmax loss~\cite{zhai2018classification}, which encourages a feature vector ${z}^m$ to be close to the prototype of the ground-truth meta-class and far away from the other meta-class prototypes.
Formally, denoting by $\mathbf{\theta}^m_{+}$ the prototype of the ground-truth, the supervised objective on a meta-class set is defined as
\begin{equation}
   \mathcal{L}_\text{meta} = -\sum_{m=1}^M \log \bigg(\frac{\exp \big( z^m\cdot\theta^{m}_{+} / \tau\big)}{\sum_{{\theta}^i \in \Theta^m} \exp \big( z^m \cdot\mathbf{\theta}^{i} / \tau\big)}\bigg),
    \label{eq:meta_loss}
\end{equation}
where each feature vector and prototype are $\ell_2$-normalized and $\tau$ represents a temperature of the softmax function.
Note that the meta-class embedding naturally introduces the inter-class relations into the model and leads better generalization for novel classes since the model learns the shared information from meta-class representations using the examples in the multiple constituent base classes.

\subsection{Unsupervised Learning of Pairwise Relations}

The combinatorial embedding is obtained by a large number of partitions through the composition of many meta-classifiers while it tends to scatter unlabeled examples over the feature space.
To learn the proper embeddings of unlabeled samples, especially the ones in novel classes, we consider two kinds of pairwise relations; one is the psuedo-label consistency between two instances and the other is the representation consistency between two augmented examples of an image.
These two objectives are learned in an unsupervised way based on the combinatorial embeddings of images as follows.

\vspace{-0.3cm}
\paragraph{Contrastive learning for pseudo-label consistency}

We capture the semantics of unlabeled data in the context of labeled ones and perform pairwise pseudo-label estimation based on similarities between two real examples.
Since the class labels in $\mathcal{D}_u$ are unknown, we provide the relational supervision for each input feature vector pair, $\left( z_i, z_j \right)$, to learn the representations properly for both labeled and unlabeled examples.
To this end, the examples with similar features are assumed to belong to the same class and regarded as a positive pair via the following procedure.

We leverage the labeled dataset $\mathcal{D}_l$ to bootstrap representations and use classification outputs from meta-classifiers $\Theta$ to infer relationships between examples.
Specifically, the positive examples are selected based on the similarities of combinatorial embedding vectors between each unlabeled example and the rest of the images in a batch, which is given by
\begin{equation}
\mathcal{P}_{z} = \{ \tilde{z} | \tilde{z} \in  B_l \cup B_u, \pi(z;\Theta) \cdot \pi(\tilde{z};\Theta) \geq \gamma)\},
\label{eq:pseudo-label}
\end{equation}
where $\mathcal{B}_l$ and $\mathcal{B}_u$ are sets of feature vectors corresponding to labeled and unlabeled examples in the current mini-batch.
Since unlabeled examples in known classes typically yield good representations thanks to labeled counterparts in the same class and the novel classes can be embedded properly in the combinatorial feature space, we expect the pseudo-label estimation by \eqref{eq:pseudo-label} is sufficiently reliable in practice under a reasonable choice of the threshold, $\gamma$.

Once the positive pairs are identified, we employ a contrastive loss~\cite{khosla2020supervised} to enforce the similarity of the positive pairs as
\begin{align}
 \mathcal{L}_\text{sim} (z) =  - \frac{1}{|\mathcal{P}_{z}|} \sum_{z_{+} \in \mathcal{P}_{z}} \log \frac{\exp\big(z \cdot \pi(z_{+};\Theta) \big)}{\sum_{\tilde{z} \in  \mathcal{B}_l \cup \mathcal{B}_u} \exp\big(z \cdot \pi(\tilde{z};\Theta)\big)},
\label{eq:pairwise_loss}
\end{align}
where $z$ and $\pi(\cdot;\Theta)$ are also $\ell_2$-normalized.
This loss term facilitates clustering novel class examples based on the cosine similarity while maintaining the representations of known class data given by~\eqref{eq:meta_loss}.
It also allows us to jointly learn the deep feature representations in both the original space and the combinatorial embedding space.

\begin{table*}[!t]
	\centering
	\caption{The mean Average Precision (mAP) for different bit-lengths on CIFAR-10 and NUS-WIDE.
	The best mAP scores are in bold.  
	GPQ with an asterisk (*) presents the result from our reproduction with the implementation provided by the original authors.}
	\label{tab:hashing}
    \scalebox{0.9}{
		\begin{tabular}{clcccccc}
		\toprule
			\multirow{2}{*}{Supervision}&
			\multicolumn{1}{c}{\multirow{2}{*}{Method}}&
			\multicolumn{3}{c}{CIFAR-10}&
			\multicolumn{3}{c}{NUS-WIDE}\\ 
                        & & 12 bits & 24 bits  & 48 bits & 12 bits & 24 bits  & 48 bits \\ \midrule
            \multirow{3}{*}{Unsupervised}& OPQ~\cite{ge2013optimized}            & 0.107   & 0.119     & 0.138   & 0.341   & 0.358     & 0.373   \\
          &  LOPQ~\cite{kalantidis2014locally}       & 0.134   & 0.127      & 0.124   & 0.416   & 0.386      & 0.379  \\
          &  ITQ~\cite{gong2012iterative}            & 0.157   & 0.165      & 0.201   & 0.488   & 0.493      & 0.503   \\ \midrule
	   \multirow{3}{*}{Supervised} & SDH~\cite{shen2015supervised}        & 0.185   & 0.193      & 0.213   & 0.471   & 0.490      & 0.507   \\ 
	  & CNNH~\cite{xia2014supervised}         & 0.210   & 0.225      & 0.231   & 0.445   & 0.463      & 0.477  \\
       &     NINH~\cite{lai2015simultaneous}       & 0.241   & 0.249     & 0.272   & 0.484   & 0.483     & 0.487   \\ \midrule
       \multirow{5}{*}{\begin{tabular}[c]{@{}c@{}}Supervised +\\ Unlabeled data\end{tabular}}  &   SSDH~\cite{zhang2017ssdh}             & 0.285   & 0.291      & 0.325   & 0.510   & 0.533    & 0.551  \\ 
       &     SSGAH~\cite{wang2018semi}              & 0.309   & 0.323      & 0.339   & 0.539   & 0.553     & 0.579  \\
    &        GPQ*~\cite{jang2020generalized}       & 0.274   & 0.290      & 0.313   & 0.598   & 0.609      & 0.615  \\
     &       SSAH~\cite{jin2020ssah}               & 0.338    & 0.370    & 0.379   & 0.569   & 0.571      &  0.596 \\ 
      &      CombEmb (ours)                & {\bf 0.667}   & {\bf 0.692}    & {\bf  0.720}  &  {\bf 0.687}  &  {\bf  0.693}   &  {\bf 0.706} \\  \bottomrule                                                                                       
		\end{tabular}}
\end{table*}

\paragraph{Consistency regularization of combinatorial embedding}

Besides the label consistency between two different examples, we perform the consistency regularization with both labeled and unlabeled data to robustify the representations obtained by combinatorial embedding in the presence of novel classes.
Given two feature vectors $z$ and $z^{\prime}$ for the two augmented views of an image $x \in \mathcal{D}_l \cup \mathcal{D}_u$, we minimize the negative cosine similarity between their combinatorial embeddings as
\begin{equation}
\mathcal{L}_\text{cons}\left(z, z^{\prime}\right) = - \frac{h(\pi(z;\Theta))}{\left\|h(\pi(z;\Theta))\right\|_{2}} \cdot \frac{\pi(z^{\prime};\Theta)}{\left\|\pi(z^{\prime};\Theta)\right\|_{2}},
\label{eq:consistency_loss}
\end{equation}
where $h(\cdot)$ denotes a prediction head~\cite{grill2020bootstrap}, and $\|\cdot\|_{2}$ denotes $\ell_2$-norm.
Following \cite{chen2021exploring}, we do not backpropagate through $\pi(z^{\prime};\Theta)$.
This loss encourages the examples of both seen and unseen classes to be embedded in the proper locations within the common embedding space, which improves the reliability of the positive pair estimation.

\subsection{Loss}
\label{sub:loss}
The total loss is a weighted sum of the three objective functions as
\begin{equation}
\mathcal{L} = \mathcal{L}_\text{meta}  + \alpha  \mathcal{L}_\text{sim}+ \beta \mathcal{L}_\text{cons},
\label{eq:overall_objective}
\end{equation}
where $\alpha$ and $\beta$ control the relative importance of the individual terms.
The proposed framework jointly performs a supervised classification and two unsupervised pairwise relation learnings. 
The learned representations based on the proposed loss function should be effective for the examples in both known and novel classes.

\subsection{Discussion}
\label{sub:discussion}
The proposed algorithm provides a unique formulation for the representation learning of novel classes, which is given by the combination of meta-classifiers learned with the examples in known labels.
The use of coarse-grained classifiers is helpful to capture common attributes across known and unknown classes and the embeddings of the examples in novel classes.

Our formulation is related to the concept of product quantization (PQ)~\cite{ge2013optimized, jegou2010product} as discussed earlier.
However, PQ is originally proposed for unsupervised hashing, which simply maximizes the variances of data in multiple subspaces and enhances retrieval performance in terms of accuracy and speed.
Its extensions to supervised learning are limited to handling known classes only~\cite{klein2019end} or fail to exploit the label information effectively for learning the representations of unlabeled novel class examples~\cite{jang2020generalized}.


\section{Experiments}
\label{sec:experiments} 
This section presents the experimental results and the characteristics of our method in the applications of image retrieval and novel class discovery given a database composed of both known and novel classes.

\subsection{Image Retrieval with Novel Class Examples}
\label{sub:retrieval}

Image retrieval is the task to identify images that belong to the same class given a query, where the database contains the examples in both known and novel classes.
In our scenarios, query images are sampled from novel classes.

\vspace{-2mm}
\paragraph{Image retrieval using combinatorial embedding}
We discuss an asymmetric search algorithm for image retrieval based on combinatorial embedding.
Let $z_q$ and $z_b$ be the feature vectors of a query image $x_q$ and a database image $x_b$, respectively.
The proposed model, which is based on $M$ partitions with $K_m$ meta-classes per partition, requires $\sum_{m=1}^{M}\log_2 (K_m)$ bits to store the approximate representation of the database image $x_b$, denoted by $\bar{z}_b = \big[{\Theta}^1[c^1_{z_b^1}], \dots, {\Theta}^M[c^M_{z_b^M}] \big]$, where $c^m_{z_b^m} \in \mathcal{C}^m$ indicates the meta-class label of $z_b^m$.
The distance between input query image and database image for asymmetric search is computed by the combination of the representations with $M$ partitions, which is given by
\begin{equation}
\sum_{m=1}^{M} \text{dist} (z_q^m, \bar{z}_b^m).
\end{equation}
where $\text{dist} (\cdot, \cdot)$ is the cosine distance function and $\bar{z}_b^m$ is the matching meta-class representation of the $m^\text{th}$ partition.

\vspace{-2mm}
\paragraph{Datasets}
We conduct experiments on four popular image retrieval benchmarks, CIFAR-10~\cite{krizhevsky2009learning}, CIFAR-100~\cite{krizhevsky2009learning}, NUS-WIDE~\cite{chua2009nus}, and CUB-200~\cite{welinder2010caltech}.
For NUS-WIDE, we use the images associated with the 21 most frequent concepts, following \cite{liu2011hashing}.
To simulate an open-set environment in the datasets, we split the classes into two subsets, known (75\%) and novel (25\%) classes, and set the half of the examples in known classes as labeled, which is identical to the protocol in~\cite{sablayrolles2017should}.
Specifically, 7, 15, 75, and 150 known classes are included in the labeled training datasets of CIFAR-10, NUS-WIDE, CIFAR-100, and CUB-200 respectively.
Note that a training dataset contains unlabeled data, which may belong to either known or novel classes.

\vspace{-2mm}
\paragraph{Baselines}
We compare the proposed approach, referred to as combinatorial embedding (CombEmb), with several image retrieval baselines based on hashing, which include OPQ~\cite{ge2013optimized}, LOPQ~\cite{kalantidis2014locally}, and ITQ~\cite{gong2012iterative}.
We also compare three supervised hashing techniques including CNNH~\cite{xia2014supervised}, NINH~\cite{lai2015simultaneous}, and SDH~\cite{shen2015supervised}, and four supervised hashing methods with additional unlabeled data such as SSDH~\cite{zhang2017ssdh}, SSGAH~\cite{wang2018semi}, GPQ~\cite{jang2020generalized}, and SSAH~\cite{jin2020ssah}. 
We extract feature descriptors from AlexNet~\cite{krizhevsky2012imagenet} pretrained on ImageNet~\cite{deng2009imagenet} for all the methods except GPQ~\cite{jang2020generalized}, which adopts the modified VGG network for CIFAR-10/100 and AlexNet for NUS-WIDE as feature extractors.

\vspace{-2mm}
\paragraph{Evaluation protocol}
Image retrieval performance is measured by the mean Average Precision (mAP).
Since all compared methods are based on hashing, their capacities are expressed by bit-lengths; the capacity of CombEmb can be computed easily using the number of meta-classifiers and the number of meta-classes.
We test three different bit-lengths $\{12, 24, 48\}$, and final results are given by the average of 4 different class splits.

\begin{table}[t!]
	\centering
	\caption{mAP scores on CIFAR-100 and CUB-200 with different number of bits.} 
	\label{tab:large_label_space}
\scalebox{0.9}{
		\begin{tabular}{clccc}
			\toprule
			Dataset & \multicolumn{1}{c}{Method}& 24 bits & 48 bits & 72 bits \\
			\midrule
			\multirow{2}{*}{CIFAR-100} & GPQ & 0.108 & 0.120 &{0.112} \\
			& CombEmb (ours) & {\bf0.154} & {\bf0.188} & {\bf0.208} \\
			\midrule
			\multirow{2}{*}{CUB-200} & GPQ & {0.167}  & {0.184} &{0.192}  \\
			& CombEmb (ours) & {\bf 0.304} & {\bf 0.337} & {\bf 0.336} \\ \bottomrule
					\end{tabular}}
\end{table}

\begin{table}[t]
	\centering
	\caption{mAP scores on CIFAR-10 and CIFAR-100 with 50\% of seen classes and another 50\% of novel ones.}
	\label{tab:lower_seen_rate}
 \scalebox{0.9}{
		\begin{tabular}{clccc}
		\toprule
			Dataset & \multicolumn{1}{c}{Method}& 24 bits & 48 bits \\
			\midrule
			\multirow{2}{*}{CIFAR-10} & GPQ &  {0.231}   &  {0.245} \\
			& CombEmb (ours) &  {\bf 0.448}   & {\bf 0.491}   \\
			\midrule
			\multirow{2}{*}{CIFAR-100} & GPQ &  {0.104}  & {0.117}    \\
			& CombEmb (ours) &  {\bf 0.165} & {\bf 0.179}  \\ \bottomrule
		\end{tabular}}
\end{table}


\vspace{-2mm}
\paragraph{Implementation details}
The backbone models and the embedding heads are fine-tuned by AdamW~\cite{loshchilov2019decoupled} with a weight decay factor of $1 \times 10^{-4}$.
For meta-classifiers, the number of meta-classes in each meta-class set ($K_m$) is fixed to 4 to simplify the experiment.
The number of meta-classifiers, $M$, is adjusted to match the bit-length of compared methods, and the dimensionality $d_2$ of $z^m$ is set to 12.
For meta-class set configuration, we generate $M$ meta-class sets by iteratively performing $k$-means clustering ($k = K_m$) over class embeddings.
We obtain the class embeddings from the classification weight vectors pretrained on labeled data. 
To ensure diverse meta-class sets, we randomly sample $Q (\ll d_1)$-dimensional subspaces of the class embeddings for each meta-class set generation.
We list the implementation details of the proposed method and the compared algorithms in the supplementary document.

\vspace{-2mm}
\paragraph{Evaluation on benchmark datasets}
We first present the performance of the proposed approach, CombEmb, on CIFAR-10 and NUS-WIDE, in comparison to existing hashing-based methods.
\cref{tab:hashing}~shows mAPs of all algorithms for three different bit-lengths, where the results of GPQ are from the reproduction on our data splits.
CombEmb achieves state-of-the-art performance in all cases on both datasets by significant margins.
This is partly because, unlike previous hashing-based approaches that suffer from limited usage of unlabeled data other than quantization error reduction or consistency regularization, our model learns discriminative representations of unlabeled examples in novel classes by utilizing their inter-class relationships with labeled data through the combination of diverse meta-classifiers.
In addition, the proposed unsupervised pairwise relation learning further improves our embedding network via enforcing similarities between unlabeled examples and their pseudo-positives.
The larger number of bits is effective for capturing the semantics in input images and achieving better performances in general.

We also apply CombEmb to more challenging datasets, CIFAR-100 and CUB-200, which contain fewer examples per class and potentially have troubles learning inter-class relations between seen and unseen classes.
\cref{tab:large_label_space}~presents that CombEmb outperforms GPQ consistently although the overall accuracies of both algorithms are lower than those on CIFAR-10 and NUS-WIDE.
On the other hand,~\cref{tab:lower_seen_rate} shows that CombEmb consistently outperforms GPQ with a fewer seen classes ($50\%$) on CIFAR-10 and CIFAR-100. 
In this experiment, $K_m$ for CIFAR-10 is set to $2$ since we have only 5 seen classes in CIFAR-10.

\begin{table}[!t]
	\centering
	\caption{Performance of different pairwise pseudo-labeling methods on CIFAR-10 and NUS-WIDE.} 
	\label{tab:pairwise_ablation}
\scalebox{0.9}{
		\begin{tabular}{clccc}
			\toprule
			Dataset & \multicolumn{1}{c}{Method}& 12 bits & 24 bits & 48 bits \\
			\midrule
			\multirow{3}{*}{CIFAR-10} & $k$-means      & 0.529 & 0.593 & 0.510 \\
			& RankStats~\cite{han2020automatically}     & 0.572 & 0.635 & 0.552 \\
			& CombEmb (ours) & {\bf 0.667} & {\bf 0.692} & {\bf 0.720} \\
			\midrule
			\multirow{3}{*}{NUS-WIDE} & $k$-means  &  0.652  &  0.638  & 0.640 \\
			& RankStats~\cite{han2020automatically}  & 0.641  &  0.649  &  0.656 \\
			& CombEmb (ours) & {\bf0.687}  & {\bf 0.693}  & {\bf 0.706} \\ \bottomrule
		\end{tabular}}
\end{table}


\begin{table}[t]
    \centering
    \captionof{table}{
    Accuracy of the proposed approach  with different combinations of the loss terms.
    }
    \label{tab:loss_combi_ablation}
    \scalebox{0.9}{
    \begin{tabular}{cccccc}
		\toprule
		\multirow{2}{*}{$\mathcal{L}_{\text{meta}}$}&
		\multirow{2}{*}{$\mathcal{L}_{\text{sim}}$}&
		\multirow{2}{*}{$\mathcal{L}_{\text{cons}}$}&
		\multicolumn{3}{c}{CIFAR-10} \\
                  & & & 12 bits & 24 bits & 48 bits\\
                \midrule
                 \checkmark  &                    &                     &  0.252 & 0.253 & 0.266\\
                 \checkmark  &                     & \checkmark & 0.510 & 0.596 & 0.623\\ 
                 \checkmark  & \checkmark &                     &  {\bf 0.687} & 0.676  &  0.619 \\ 
                 \checkmark  & \checkmark & \checkmark  &  0.667  & {\bf 0.692} & {\bf 0.720} \\ \bottomrule
	\end{tabular}
    }
\end{table}

\vspace{-2mm}
\paragraph{Analysis on pairwise label estimation}

To understand the effectiveness of the positive pair estimation proposed in~\eqref{eq:pseudo-label}, we compare the strategy with the following two baselines: 1) using $k$-means clustering on the feature vectors to assign labels of unlabeled data ($k$-means), and
2) adopting rank statistics~\cite{han2020automatically} between feature vectors in the original space to estimate pairwise labels (RankStats).
For the first baseline, we assume the ideal case in which the number of clusters is known and equal to the exact number of classes appearing in training.
\cref{tab:pairwise_ablation}~implies that our label estimation strategy based on combinatorial embeddings outperforms other baselines.

\vspace{-2mm}
\paragraph{Analysis of loss functions}

\cref{tab:loss_combi_ablation}~demonstrates the contribution of individual loss terms on CIFAR-10.
Each of the three loss terms, especially the similarity loss ($\mathcal{L}_{\text{sim}}$) and consistency loss ($\mathcal{L}_{\text{cons}}$), turn out to be effective for improving accuracy consistently.
Also, the similarity loss together with the consistency loss is helpful to obtain the desirable tendency in accuracy with respect to bit-lengths.
Note that the proposed combinatorial embedding learns basic representations suitable for both seen and unseen examples and the unsupervised pairwise relation learning improves performance dramatically on top of them. 

\vspace{-2mm}
\paragraph{Analysis on the number of meta-classes}
\cref{tab:km_ablation} presents the performance of CombEmb by varying the number of the meta-classes $K_m$ while the bit-lengths are controlled as the same values by adjusting $M$, the number of meta-classifiers.
We observe that larger bit-lengths are consistently helpful for improving the accuracy of CombEmb while $K_m$ and $M$ alone have limited impacts on performance.


\begin{table}[t]
    \centering
    \captionof{table}{Sensitivity of CombEmb in mAP to the number of meta-classes $K_m$, where the bit-length is controlled by adjusting the number of meta-classifiers $M$ given $K_m$.}
    \label{tab:km_ablation}
    \scalebox{0.9}{
		\begin{tabular}{ccccccc}
			\toprule
			\multirow{2}{*}{$K_m$} & \multicolumn{3}{c}{CIFAR-100}& \multicolumn{3}{c}{CUB200} \\
			 & 24 bits & 48 bits & 72 bits & 24 bits & 48 bits & 72 bits \\
			\midrule
			2  &  0.153    &  0.175 & 0.179 & 0.296& 0.330& 0.334\\
			4  &  0.154    &  0.188 & 0.208 & 0.304& 0.337&0.336\\
			8 &  0.141   &  0.196  & 0.223 & 0.291& 0.337& 0.338\\ \bottomrule
			\end{tabular}}
\end{table}

\begin{table}[!t]
    \centering
    \captionof{table}{mAP scores on CIFAR-10 with fewer labeled examples of seen classes, where $7$ classes are set as seen classes.}
    \label{tab:labeled_rate_ablation}
    \vspace{1mm}
    \scalebox{0.9}{
		\begin{tabular}{clccc}
			\toprule
			Ratio & \multicolumn{1}{c}{Method}& 12 bits & 24 bits & 48 bits \\
			\midrule
			\multirow{2}{*}{30\% labeled} & GPQ  &  0.217    &  0.207 & 0.223 \\
			& CombEmb (ours)&  {\bf 0.354}   & {\bf 0.572}  & {\bf 0.697} \\
			\midrule
			\multirow{2}{*}{10\% labeled} & GPQ & 0.177    &  0.190 &  0.191  \\
			& CombEmb (ours) & {\bf 0.184}  & {\bf 0.264} & {\bf 0.498}		\\ \bottomrule
			\end{tabular}}
\end{table}


\begin{table*}[!t]
    \centering
    \caption{Comparison with novel class discovery methods on CIFAR-10, CIFAR-100, and Tiny-ImageNet in terms of ACC, NMI, and ARI.
Dagger ($\dagger$) denotes that the dimensionality of the classifier in the original method is extended to the total number of classes in the dataset.}
    \label{tab:novel_discovery_cifar10}
    \vspace{0.1cm}
    \scalebox{0.9}{
		\begin{tabular}{clccccccccc}
			\toprule
			\multicolumn{1}{c}{\multirow{2}{*}{Dataset}} &
			\multicolumn{1}{c}{\multirow{2}{*}{Method}}&
			\multicolumn{3}{c}{ACC}&
			\multicolumn{3}{c}{NMI}&
			\multicolumn{3}{c}{ARI}\\
                          & & Seen & \hspace{-0.1cm}Unseen\hspace{-0.1cm} & Total & Seen & \hspace{-0.1cm}Unseen\hspace{-0.1cm} & Total & Seen & \hspace{-0.1cm}Unseen\hspace{-0.1cm} & Total \\
\midrule
           \multirow{7}{*}{CIFAR-10} & DTC$^\dagger$~\cite{Han2019learning}                    & 77.44 & 72.00  & 75.81 & 62.91 & 65.82 & 63.25 & 61.98 & 65.33 & 56.89 \\                        
            & RankStats$^\dagger$~\cite{han2020automatically}            & 82.22 & 88.23 & 84.01 & 72.03 & 75.17 & 73.65 & 66.68 & 79.54 & 67.08 \\ 
            & NCL$^\dagger$~\cite{zhong2021neighborhood}                & 78.47 & 85.23 & 80.50 & 69.09 & 74.97 & 66.18 & 68.06 & 78.35 & 56.56 \\ 
            & DualRank$^\dagger$~\cite{zhao2021novel}                       & 83.89 & 86.91 & 84.79 & 72.97 & 74.37 & 74.98 & 67.61 & 78.63 & 68.02 \\ \cmidrule(lr){2-11}
            & ORCA~\cite{cao2021open}                               & 87.70 & 92.83 & 89.24 & 74.34 & {\bf81.81} & 78.21  & 76.46  & {\bf88.13}  & 78.12 \\ 
            & GCD~\cite{vaze2022generalized}                     & 88.82 & 88.23 & 88.63 &76.28 & 75.53 & 77.99 & 77.52  & 81.19  & 76.75 \\ 
            & CombEmb (ours)                              		      & {\bf89.02} & {\bf92.96}  & {\bf89.98} & {\bf77.97} & {80.43}  &  {\bf79.83} & {\bf80.26} &{85.63} & {\bf79.19}\\                                                                   
                         \midrule
            \multirow{7}{*}{CIFAR-100} & DTC$^\dagger$~\cite{Han2019learning}                    & 42.67 & 27.44 & 38.86 & 54.71 & 44.77 & 50.04 & 20.27 & 23.84 & 15.24 \\                        
            & RankStats$^\dagger$~\cite{han2020automatically}             & 47.33 & 34.79 & 42.49 & 62.80 & 50.56  & 58.29 & 30.93 & 21.05 & 22.63 \\ 
            & NCL$^\dagger$~\cite{zhong2021neighborhood}                 & 53.13  & 35.80 & 48.80 & 63.23 & 54.24  & 58.56 & 38.93 & 27.68  &  30.06\\ 
            & DualRank$^\dagger$~\cite{zhao2021novel}                        & 46.08 & 36.47 & 42.54 & 61.59 & 52.20 & 57.57 & 29.83 & 24.09 & 23.00 \\ \cmidrule(lr){2-11}
            & ORCA~\cite{cao2021open}                                & 64.85 & 44.83 & 56.77 & 66.18 & 58.89 & 62.64  & 46.22  & 38.68  & 38.06 \\ 
            & GCD~\cite{vaze2022generalized}                      & 66.04 & 38.53 & 55.59 & 69.61 & 57.67 & 64.56 & 49.99 & 30.38 & 38.85 \\ 
            & CombEmb (ours)                              		      & {\bf69.19} & {\bf51.41}  & {\bf62.11} & {\bf70.77} &  {\bf63.81} & {\bf67.71} &{\bf52.22} & {\bf43.37} & {\bf43.37}\\                                                                   
                         \midrule
            \multirow{7}{*}{\makecell{Tiny-ImageNet}} & DTC$^\dagger$~\cite{Han2019learning}                    & 16.76 & 14.47 & 16.19 & 34.41 & 32.70 & 32.19 & \ \ 7.23 & \ \ 7.72   & \ \ 5.72 \\                        
            & RankStats$^\dagger$~\cite{han2020automatically}            & 31.49 & 19.75 & 26.87 & 54.76 & {47.52}  & 51.14 & 16.16 & \ \ 9.86 & 11.87 \\ 
            & NCL$^\dagger$~\cite{zhong2021neighborhood}                & 35.27 & 18.90 & 31.18 & 55.08 & 47.16 & 51.10 & 17.89 & 10.01 & 12.85 \\ 
            & DualRank$^\dagger$~\cite{zhao2021novel}                       & 29.76 & 18.93 & 26.70 & 53.86 & 48.03 & 50.53 & 14.88 & \ \ 9.69  & 11.08 \\ \cmidrule(lr){2-11}
            & ORCA~\cite{cao2021open}                               & 47.46 & 22.55 & 38.58  & 60.26 & 49.08 & 55.23  & 27.62  & 13.37  & 19.72 \\ 
            & GCD~\cite{vaze2022generalized}                     & 45.77 & 20.59 & 38.03 & 61.58 & 49.74 & 56.54 & 27.21  & 11.45  & 19.32 \\ 
            & CombEmb (ours)                              		      & {\bf53.76} & {\bf27.41}  & {\bf43.73} & {\bf63.91} & {\bf54.07} & {\bf59.01} &{\bf33.43} & {\bf17.70} & {\bf23.40}\\           \bottomrule
		\end{tabular}}
\end{table*}

\vspace{-2mm}
\paragraph{Results with fewer labeled data}
We perform experiments when $30\%$ and $10\%$ of the examples in the seen classes are labeled and present the results in~\cref{tab:labeled_rate_ablation}.
These settings are more realistic and challenging than the environment of our main experiments.
Although the overall accuracy is degraded compared to the main results due to the lack of supervision, the proposed algorithm outperforms GPQ by large margins regardless of bit-lengths.
Note that when we use large bit-lengths (48 bits), the performance gap becomes more significant than the experiments in~\cref{tab:hashing}, achieving $3.1\times$ and $2.6\times$ accuracy gains with $30\%$ and $10\%$ of labeled seen-class examples, respectively.


\subsection{Categorization with Novel Class Discovery}
\label{sub:novel_class_discovery}

We evaluate the performance of CombEmb on image categorization with novel class discovery.
The goal of this task is to cluster unlabeled examples that belong to either seen or novel classes into a predefined number of groups based on their semantic relations.
This task is more natural and challenging than the standard novel class discovery that only considers unseen classes in unlabeled data.

\vspace{-2mm}
\paragraph{Datasets}
We evaluate the proposed approaches on three standard datasets including CIFAR-10, CIFAR-100, and Tiny-ImageNet.
Similar to the experiments for image retrieval, we split the classes into 75\% seen and 25\% novel classes: the first 7, 75, and 150 classes in CIFAR-10, CIFAR-100, and Tiny-ImageNet are respectively selected as seen classes.
Following~\cite{cao2021open,vaze2022generalized}, we assume that the half of examples in seen classes are labeled while setting the rest in seen classes and examples in novel classes as unlabeled.
Note that the unlabeled data may belong to either known or novel classes.%

\begin{figure*}[t]
\centering
\begin{subfigure}[b]{1.0\linewidth}
\centering
\includegraphics[width=0.8\linewidth]{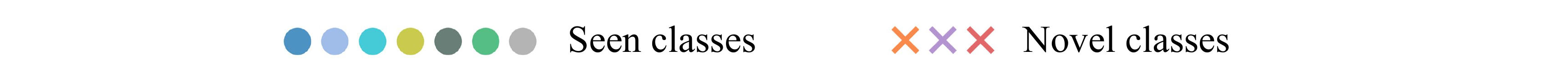}
\end{subfigure}
\begin{subfigure}[b]{0.19\linewidth}
\includegraphics[width=1\linewidth]{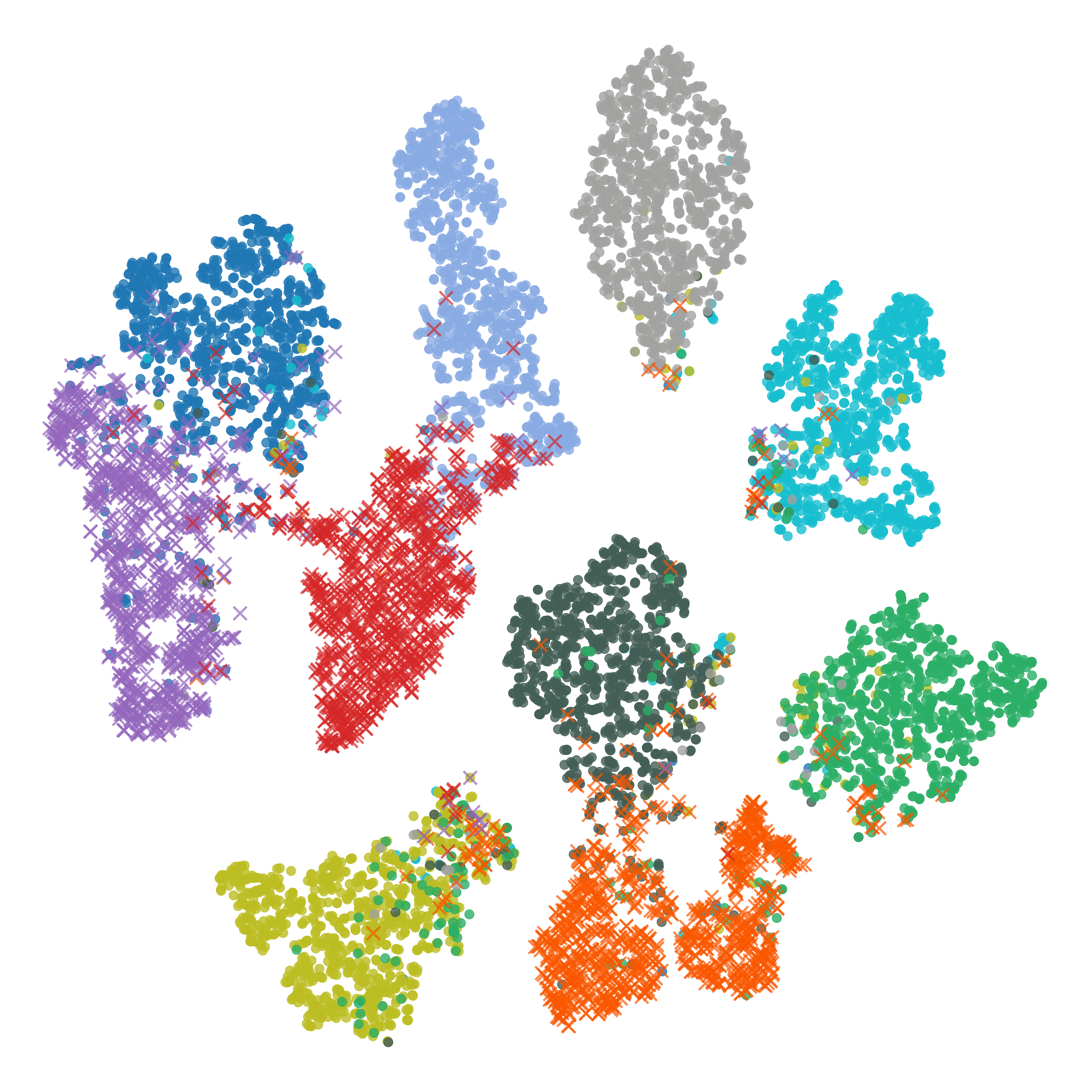}
\caption{RankStats}
\end{subfigure}
\hspace{0.03cm}
\begin{subfigure}[b]{0.19\linewidth}
\includegraphics[width=1\linewidth]{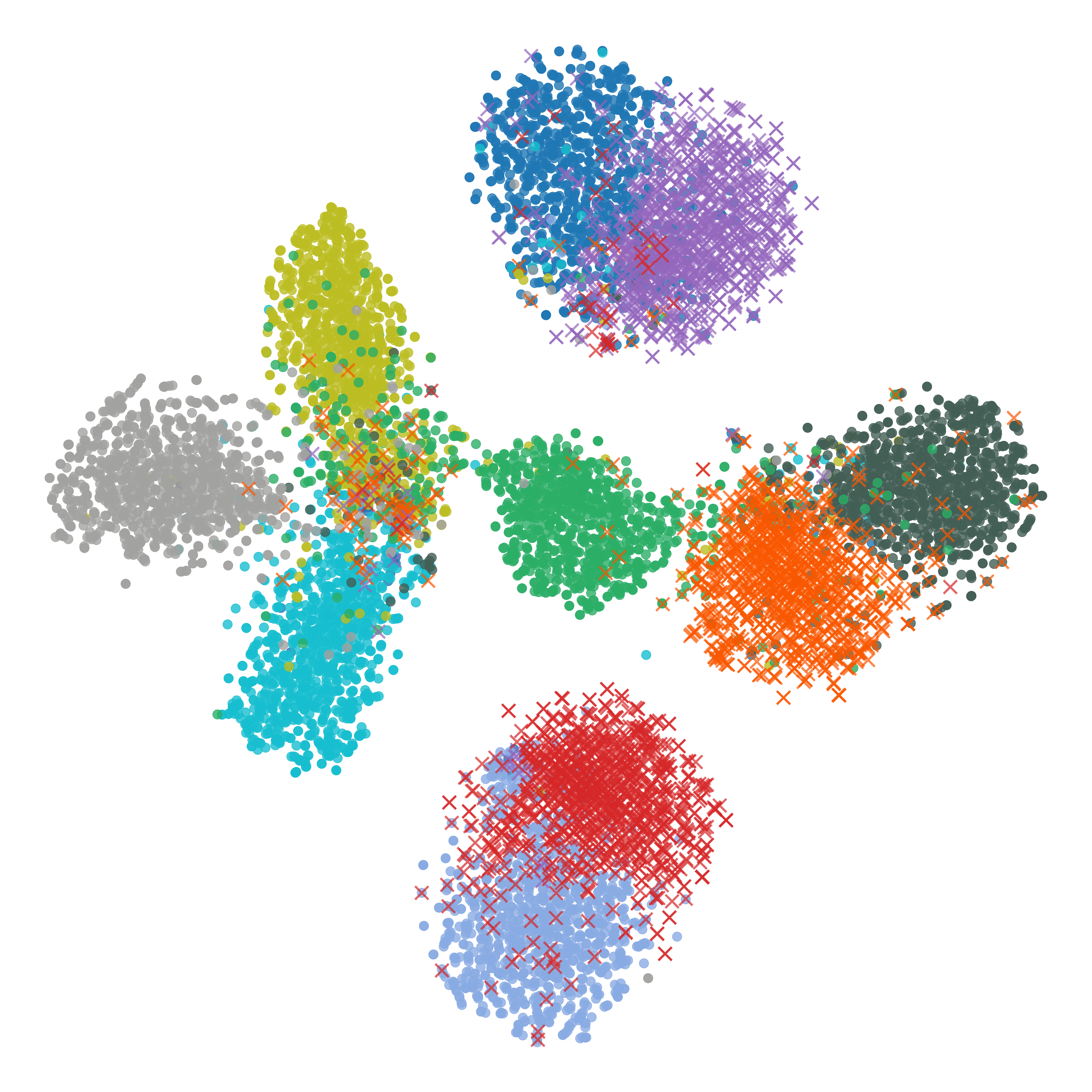}
\caption{NCL}
\end{subfigure}
\hspace{0.03cm}
\begin{subfigure}[b]{0.19\linewidth}
\includegraphics[width=1\linewidth]{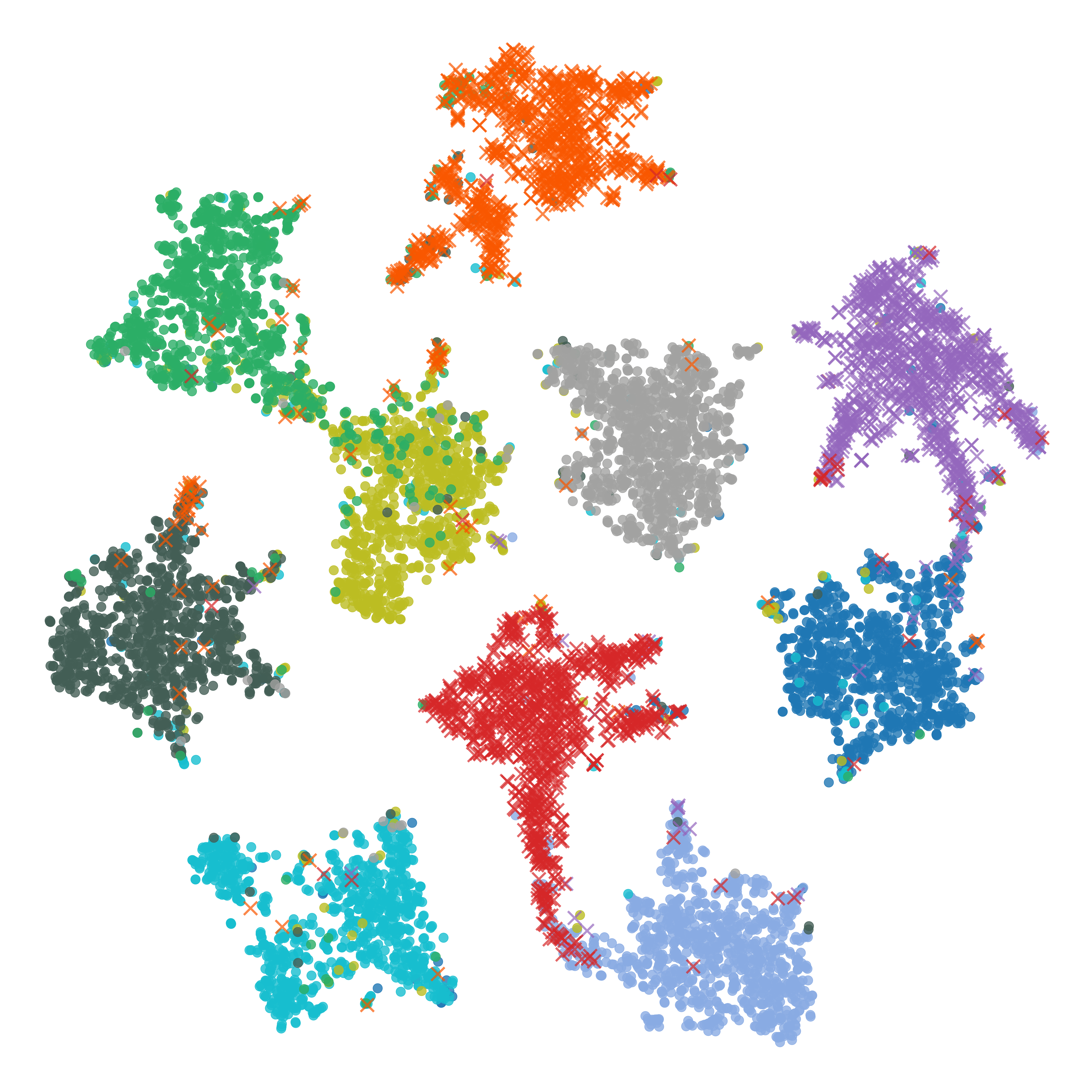}
\caption{ORCA}
\end{subfigure}
\hspace{0.03cm}
\begin{subfigure}[b]{0.19\linewidth}
\includegraphics[width=1\linewidth]{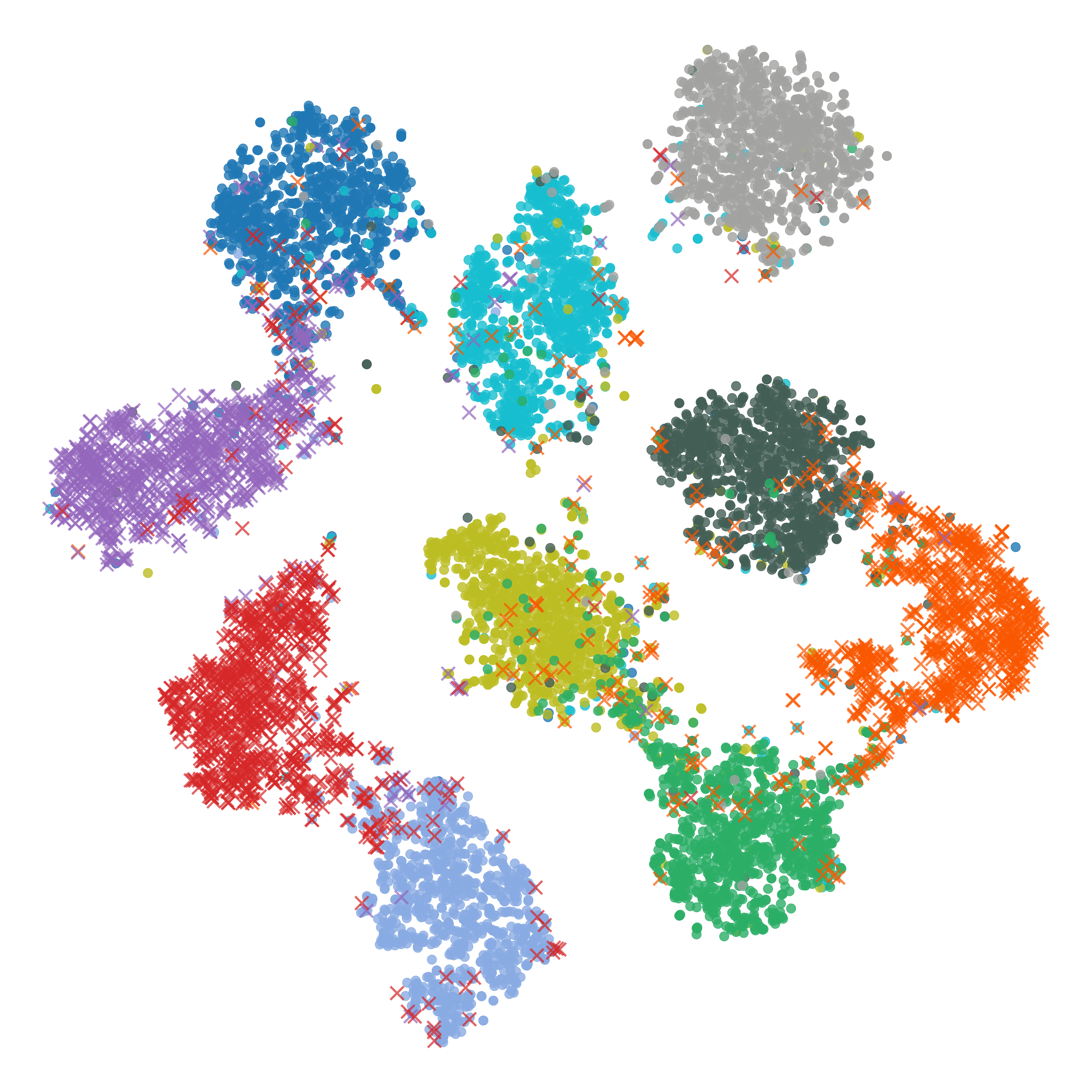}
\caption{GCD}
\end{subfigure}
\hspace{0.03cm}
\begin{subfigure}[b]{0.19\linewidth}
\includegraphics[width=1\linewidth]{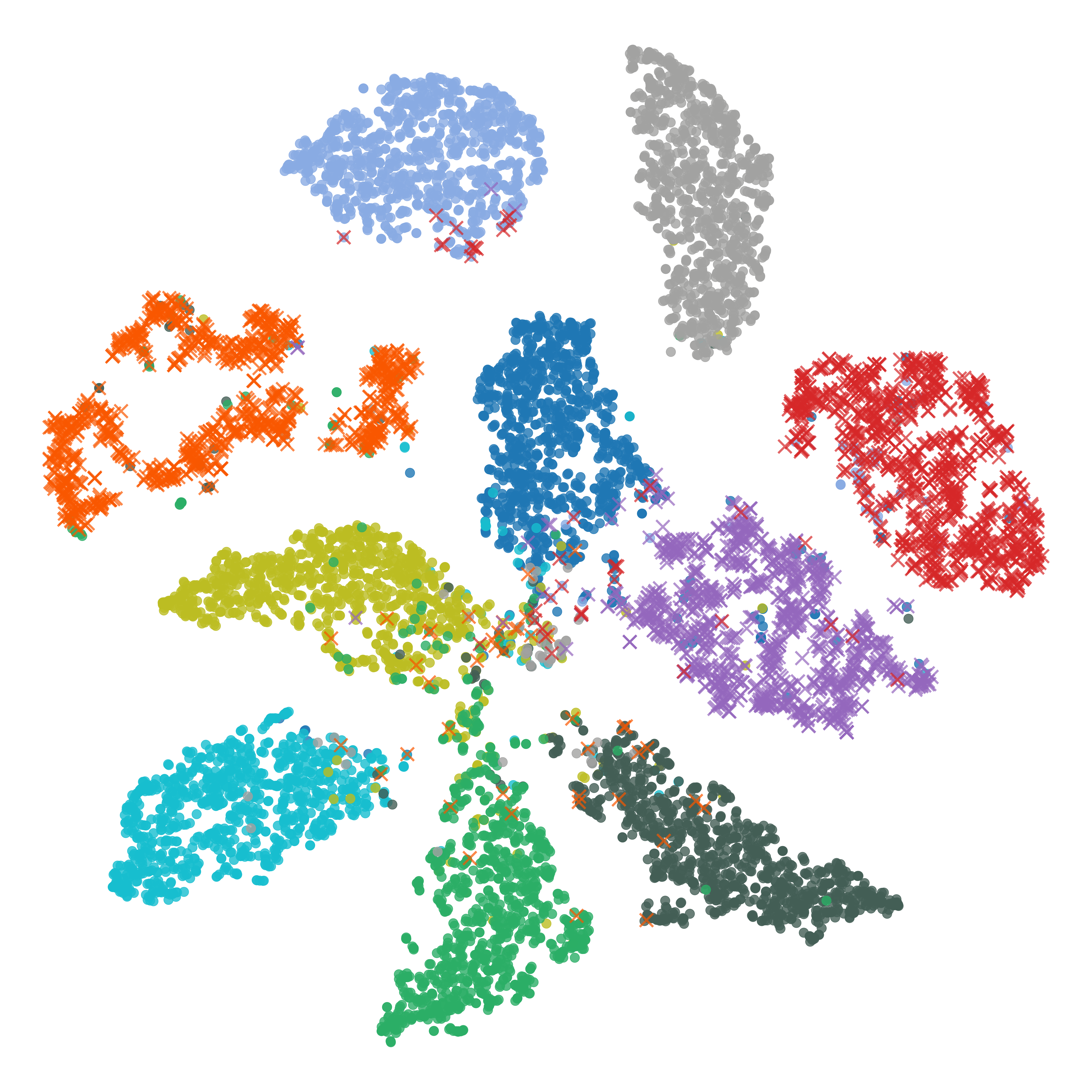}
\caption{CombEmb (ours)}
\end{subfigure}
\caption{t-SNE visualization for the data embedding of CIFAR-10, learned by Rankstats, NCL, ORCA, GCD, and CombEmb.
The visualization is based on 7 seen classes and 3 novel classes. 
Colors represent their ground-truth labels.}
\label{fig:discovery_tsne}
\end{figure*}

\vspace{-2mm}
\paragraph{Baselines}
We compare CombEmb with the state-of-the-art approaches in novel class discovery including DTC~\cite{Han2019learning}, RankStats~\cite{han2020automatically}, NCL~\cite{zhong2021neighborhood}, and DualRank~\cite{zhao2021novel}. 
We additionally consider two more methods in a similar setting: ORCA~\cite{cao2021open} and GCD~\cite{vaze2022generalized}.
Since the classification heads for unlabeled data in DTC, RankStats, NCL, and DualRank cannot handle seen classes, we increase the dimensionality of the classifiers to the total number of classes in datasets.
This extension requires estimating the number of novel classes~\cite{Han2019learning,vaze2022generalized} unless it is given in advance.
For all datasets, we use ResNet-18 as a backbone and pretrain all the compared methods with SimCLR~\cite{chen2020simple} while DTC, RankStats, and NCL additionally fine-tune their models with the labeled data.
We describe the implementation details of all algorithms in the supplementary document.
For evaluation, we first identify the cluster membership of each example in the test set via $k$-means clustering, and then compute clustering accuracy (ACC), Normalized Mutual Information (NMI), and Adjusted Rand Index (ARI) using the cluster indices.
Note that, to report ACC, we solve the optimal assignment problem using the Hungarian algorithm~\cite{kuhn1955hungarian}.

\vspace{-2mm}
\paragraph{Results}
\cref{tab:novel_discovery_cifar10}~presents the clustering performance of the learned representations by all the compared methods on all the three datasets.  
CombEmb outperforms the baselines for both seen classes and novel classes in most cases.
The results show that CombEmb learns effective representations for clustering in the presence of unseen classes in training datasets, which leads to state-of-the-art performance.
Figure~\ref{fig:discovery_tsne}~visualizes the embeddings learned by RankStats, NCL, ORCA, GCD, and CombEmb on CIFAR-10.
Our method embeds known and novel classes in a more discriminative way through supervised combinatorial classification followed by unsupervised learning, while other methods suffer from learning the discriminative representations, especially between the novel classes and their closest seen classes.



\section{Conclusion} 
\label{sec:conclusion}

This paper presents a novel representation learning approach, where only a subset of training examples are labeled while unlabeled examples may contain both known and novel classes.
To address this problem, we proposed a combinatorial learning framework, which identifies and localizes the examples in unseen classes using the composition of the outputs from multiple coarse-grained classifiers on heterogeneous meta-class spaces.
Our approach further improves the semantic structures and the robustness of the representations via unsupervised relation learning.
The extensive experiments on the standard benchmarks for image retrieval and image categorization with novel class discovery demonstrate the effectiveness of the proposed algorithm, and the various ablative studies show the robustness of our approach.


\vspace{-0.2cm}
\paragraph{Acknowledgments}
\label{Acknowledgments}


This work was partly supported by Samsung Advanced Institute of Technology (SAIT) and the NRF Korea grant [No. 2022R1A2C3012210, Knowledge Composition via Task-Distributed Federated Learning; No.2022R1A5A708390811, Trustworthy Artificial Intelligence].
It is also supported in part by the IITP grants [2021-0-02068; 2021-0-01343] funded by the Korea government (MSIT).

{\small
\bibliographystyle{ieee_fullname}
\bibliography{egbib}
}

\end{document}